\documentclass{article}
\usepackage[utf8]{inputenc} 

\usepackage{tikz}
\usetikzlibrary{matrix,chains,positioning,decorations.pathreplacing,arrows,automata,calc,scopes}
\tikzset{%
  every neuron/.style={
    circle,
    draw,
    minimum size=0.5cm
  },
  neuron missing/.style={
    draw=none,
    scale=2.5,
    text height=0.333cm,
    execute at begin node=\color{black}$\vdots$
  }
}

\usepackage{xcolor}
\usepackage{amsmath}
\usepackage{mathtools}
\usepackage[T1]{fontenc}    
\usepackage{hyperref}       
\usepackage{url}            
\usepackage{booktabs}       
\usepackage{amsfonts}       
\usepackage{nicefrac}       
\usepackage{microtype}      
\usepackage{graphicx}
\usepackage{lmodern}
\usepackage{subcaption}
\usepackage{cleveref}       
\usepackage{xcolor}
\usepackage{amsmath}
\usepackage{amssymb}
\usepackage[T1]{fontenc}    
\usepackage{hyperref}       
\usepackage{url}            
\usepackage{booktabs}       
\usepackage{amsfonts}       
\usepackage{nicefrac}       
\usepackage{microtype}      
\usepackage{graphicx}
\usepackage{subcaption}
\usepackage{cleveref}       

\usepackage{natbib}
\usepackage{doi}
\usepackage{natbib}
\usepackage{doi}

\usepackage{lineno}         
\usepackage{authblk}        
\usepackage{subcaption}

\usepackage{listings}
\usepackage{listings-options}
\usepackage{color}

\hypersetup{colorlinks = true,
linkcolor = purple,
urlcolor  = blue,
citecolor = cyan,
anchorcolor = black}

\newcommand{\ud}{\mathrm{d}}

\title{Beyond Rate Coding: Surrogate Gradients Enable Spike Timing Learning in Spiking Neural Networks}
\author{
Ziqiao Yu$^{1}$,
Pengfei Sun$^{1}$\thanks{Corresponding author. p.sun@imperial.ac.uk},
Danyal Akarca$^{1,2,3}$,
Dan F.~M.~Goodman$^{1}$ 
\\
$^{1}$Department of Electrical and Electronic Engineering, Imperial College London, London, UK 
$^{2}$Imperial-X, Imperial College London, London, UK 
$^{3}$MRC Cognition and Brain Sciences Unit, University of Cambridge, Cambridge, UK 
\\
}

\date{}

\begin{document}
\maketitle


\begin{abstract}
	\begin{itemize}
        The surrogate gradient descent algorithm enabled spiking neural networks to be trained to carry out challenging sensory processing tasks, an important step in understanding how spikes contribute to neural computations.
        However, it is unclear the extent to which these algorithms fully explore the space of possible spiking solutions to problems.
        We investigated whether spiking networks trained with surrogate gradient descent can learn to make use of information that is only encoded in the timing and not the rate of spikes.
        We constructed synthetic datasets with a range of types of spike timing information (interspike intervals, spatio-temporal spike patterns or polychrony, and coincidence codes).
        We find that surrogate gradient descent training can extract all of these types of information.
        In more realistic speech-based datasets, both timing and rate information is present.
        We therefore constructed variants of these datasets in which all rate information is removed, and find that surrogate gradient descent can still perform well.
        We tested all networks both with and without trainable axonal delays.
        We find that delays can give a significant increase in performance, particularly for more challenging tasks.
        To determine what types of spike timing information are being used by the networks trained on the speech-based tasks, we test these networks on time-reversed spikes which perturb spatio-temporal spike patterns but leave interspike intervals and coincidence information unchanged.
        We find that when axonal delays are not used, networks perform well under time reversal, whereas networks trained with delays perform poorly.
        This suggests that spiking neural networks with delays are better able to exploit temporal structure.
        To facilitate further studies of temporal coding, we have released our modified speech-based datasets. 
	\end{itemize}
\end{abstract}

\section{Introduction}
\label{sec:introduction}

Spiking Neural Networks (SNNs) are biologically inspired models that compute via discrete, sparse spikes, rather than continuous activations \citep{maass1997networks}. This event‑driven framework not only captures rich temporal patterns (such as inter‑spike intervals and cross‑neuron synchrony) but also powers energy‑efficient neuromorphic hardware  \citep{painkras2013spinnaker, akopyan2015truenorth, davies2018loihi, orchard2021efficient,su2024ultra}.

Despite their promise, a persistent challenge remains:  how to convert precise spike timing into effective learning signals. Surrogate Gradient Descent (Surrogate GD) provides a practical solution by replacing the non‑differentiable spike with a smooth surrogate, thereby allowing backpropagation through time \citep{neftci2019surrogate}. Subsequent work has improved its efficiency and adaptiveness by enforcing sparser gradients \citep{perez2021sparse} and by developing more adaptive surrogate functions \citep{wang2023adaptive}. Surrogate GD–trained networks have matched or outperformed rate‑based recurrent neural networks on several neuromorphic benchmarks  \citep{shrestha2018slayer, 10757311, dampfhoffer2024neuromorphic, hammouamri2024learning, sun2025towards}, but it remains unclear whether these successes stem from genuine temporal coding or from reliance on coarse firing‑rate features. 

To address this ambiguity, researchers have injected various temporal inductive biases: heterogeneous time constants \citep{fang2021incorporating, perez2021neural,habashy2024adapting,yu2025neuronal}; learnable axonal delays  \citep{shrestha2022spikemax, sun2022axonal, 10181778,sun2025algorithm}; synaptic delays \citep{zhang2020supervised, yu2022improving, sun2024delay,hammouamri2024learning,queant2025delrec}; and dendritic delay models \citep{d2024denram, zheng2024temporal}. Yet even with these enhancements, there has been no systematic test of whether or not Surrogate GD finds networks that make use of precise timing.

This uncertainty leads to a central question: \textit{can Surrogate GD-trained SNNs learn and depend on spike timing codes beyond rate statistics}?
This question is important because if they were just using rate-based information, then (a) they are not making use of all the information that is available, and are therefore less efficient than they could be, (b) it limits their effectiveness as a model of spike-based computation in the brain, and (c) it raises the question of whether or not there is an advantage of using such expensive training methods compared to simply approximating rate-based networks. {Previous work has demonstrated that surrogate gradient methods can be trained to reproduce prescribed spiking patterns or low-dimensional spike manifolds~\citep{zenke2018superspike}. However, such approaches typically impose strong constraints on the input or target spike structure~\citep{zenke2021remarkable}. }
To address this central question, we used both synthetic datasets as well as more complicated real-world benchmarks. First, we construct synthetic benchmark tasks that isolate the contribution of inter-spike interval (ISI), coincidence, and cross-channel inter-spike interval (CCISI, a type of spatio-temporal spike pattern or polychrony) under controlled spike counts, building upon prior work that highlights the ability of SNNs to leverage temporal coding for classification~\citep{bohte2002error,mirsadeghi2021stidi}. Then, we evaluate models on datasets with real-world temporal complexity \citep{cramer2020heidelberg}, introducing spike count-normalized variants to remove rate confounds. We assess robustness under biologically inspired perturbations: Gaussian jitter applied per-spike and per-neuron~\citep{amarasingham2012conditional,agmon2012novel}, and spike deletions~\citep{mainen1995reliability,chen2022reliability}. Finally, we analyse the effect of time reversal that selectively disrupts spike patterns while leaving coincidences and ISIs unchanged.

Our findings show that Surrogate GD-trained SNNs can extract fine-grained temporal structure even without spike rate information, including intra-neuron ISIs, cross-neuron synchrony or coincidence, and cross-channel causal ISI. Delay learning further enhances the model’s ability to exploit such timing features, especially for long-range causal gaps that exceed the membrane integration window controlled by the time constant. Across perturbations we observe distinct degradation patterns that reveal both vulnerabilities and robustness in the model's temporal encoding. Notably, under time reversal, Surrogate GD-trained SNNs not only leverage reversal-invariant features like ISI and coincidence, but also capture causal dependencies in spike timing (as in CCISI), indicating a layered temporal representation. {The implementation details are available at \url{https://github.com/neural-reckoning/temporal-shd}.
}
\section{Surrogate Gradient Descent Enables Learning of Diverse Spike-Timing Codes}
\label{sec:sg_learns_timing}

Throughout this section we analyse three synthetic benchmarks that isolate specific temporal features---intra-neuron inter-spike intervals (ISI), inter-neuron coincidence, and cross-channel inter-spike intervals (CCISI).

\subsection{Models}

The model is a feedforward spiking neural network, consisting of an input layer of $N$ neurons, a fully connected spiking hidden layer (with 100 neurons for the ISI and CCISI tasks, see \cref{sec:sg_learns_isi,sec:sg_learns_ccisi}, and 3 neurons for the coincidence task, see \cref{sec:sg_learns_coin}), and a fully connected spiking output layer whose size equals the number of classes.
The membrane time constant $\tau$ is shared between all neurons, and learned during training.
We compare two architectures: (i) synaptic weights and $\tau$ are learnable, hereafter referred to as Learnable Tau; and (ii) the same model augmented with learnable axonal delays, referred to as Tau + Delay. 

Training is implemented in the SLAYER framework \citep{shrestha2018slayer}. In this framework, during forward propagation, the spike threshold uses a Heaviside function $H(x)$, while during back-propagation it employs the smooth surrogate
\begin{equation}
\frac{\ud H(x)}{\ud x}  \triangleq \frac{1}{(\alpha \cdot |x| + 1)^2}, \quad \alpha = 100,
\label{eq:surrogate_grad}
\end{equation}
following the general form proposed by \citep{neftci2019surrogate}. Unless otherwise noted, this surrogate gradient is used consistently across all models. 
{This surrogate derivative provides a simple mechanism by which gradients become sensitive to spike timing. When an input spike arrives slightly earlier or later, the postsynaptic membrane potential crosses threshold at a different time, and the surrogate derivative in Eq.~(\ref{eq:surrogate_grad}) produces a correspondingly shifted gradient. In combination with learned delays, this creates an effective temporal receptive field: the network strengthens weights or adjusts delays for spike timings that consistently support correct classification.}

During training, classification is supervised with a Spikemax loss on the output spikes over the simulation time \citep{shrestha2022spikemax}.
{In the synthetic CCISI and ISI experiments, both the membrane time constant~$\tau$ and the axonal delays are treated as fully learnable parameters and optimized jointly with all network weights using the surrogate gradient method \citep{neftci2019surrogate, sun2022axonal}.  }

\begin{figure*}[!htbp]
  \centering
  \includegraphics[width=\linewidth]{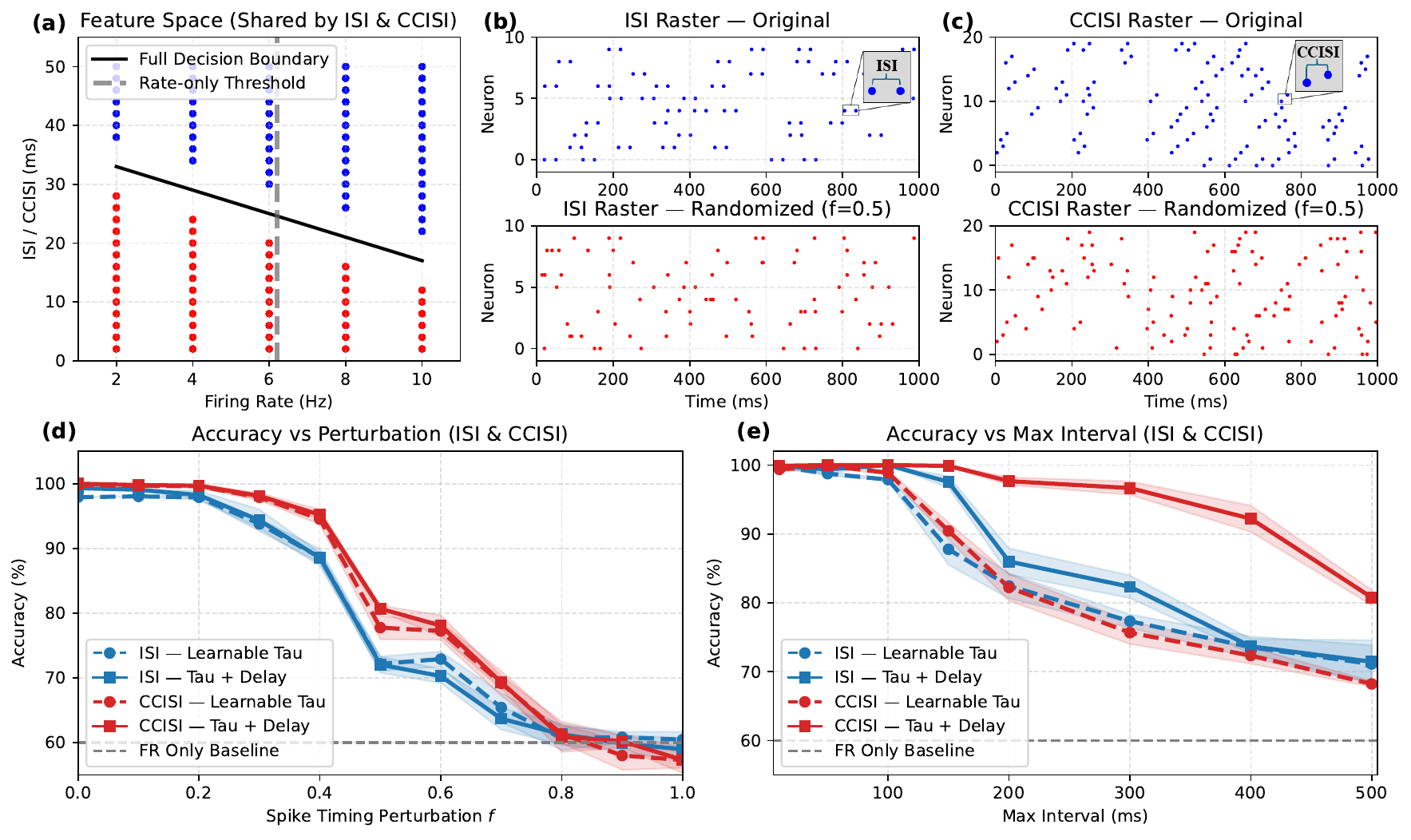}
  \caption{\textbf{ISI and CCISI-based datasets.} 
  (a) Shared feature space spanned by firing rate and ISI/CCISI interval. 
  A full decision boundary (black) separates classes more effectively than a rate-only threshold (gray). 
  (b) Example ISI spike trains with fixed intervals (top) and their randomized versions with $f=0.5$ (bottom). 
  (c) Example CCISI spike trains with fixed cross-neuron delays (top) and their randomized counterparts with $f=0.5$ (bottom). 
  (d) Test accuracy under increasing spike timing perturbation $f$ (evaluated at maximum interval of 50\,ms), comparing models with a learnable membrane constant $\tau$ and those augmented with delays. 
  (e) Accuracy across increasing maximum ISI/CCISI intervals.}
  \label{fig:isi_ccisi_result}
\end{figure*}

\subsection{Surrogate GD Learns Inter-Spike Intervals}
\label{sec:sg_learns_isi}

\paragraph{{Task setup.}}
We first evaluate whether SNNs trained with surrogate gradients can learn from fine-grained inter-spike interval (ISI) structure. We do this by constructing a dataset where it is impossible to perform much above chance level without using ISIs.

In our dataset, each sample contains $N=10$ input neurons, with each neuron generating spike pairs with a fixed class-specific ISI $\delta$ and firing rate $r$ (\cref{fig:isi_ccisi_result}(a)). Spikes are distributed within a window and placed at non-overlapping positions (\cref{fig:isi_ccisi_result}(b) upper). The number of spikes is $r \cdot \tfrac{T}{1000}$. When the max interval is less than 100 ms, we use $T=1,000$ and $r\in[1,50]$, and for longer intervals we use $T=10,000$ and $r\in[0.1,1]$ to ensure that placement of all pairs is possible.

To simulate timing degradation, a fraction $f \in [0, 1]$ of spikes in each neuron's train are removed and replaced by new spikes sampled uniformly over $[0,T)$. This procedure ensures that the total number of spikes remains unchanged and that no two spikes overlap in time for the same neuron. When $f=0$ all spikes come in pairs with an unambiguous ISI. When $f=1$, the spike train is fully randomized, removing all ISI information while still preserving the original spike count.

\paragraph{{Results.}}
In the clean case where $f=0$ the models achieve approximately 100\% accuracy (\cref{fig:isi_ccisi_result}(d)), demonstrating that surrogate gradient descent is able to extract temporal structure from spike trains encoded in the form of interspike intervals. Models with and without learnable delays perform about equally across the whole range of perturbation levels $f$, indicating that the delay module does not provide additional benefit in this setting.
As the perturbation factor $f$ increases from 0 to 1, spike timing information becomes increasingly disrupted, leading to degraded model performance, although performance remains close to perfect until $f=0.2$ (20\% of spikes disrupted) and only substantially degrades for $f\geq 0.4$. For $f > 0.5$ (\cref{fig:isi_ccisi_result}(b) lower), most ISIs are broken, as each interval requires a pair of spikes. The model then relies primarily on firing rate, with accuracy converging to approximately 60\%. This is above chance level (50\%) but reflects the limit of rate-based discrimination.

\subsection{Delay Module Enhances Cross-Channel ISI Code Learning}
\label{sec:sg_learns_ccisi}
\paragraph{{Task setup.}}
To further probe the role of delay learning, we designed a cross-channel ISI (CCISI) task in which spike timing is explicitly causal across neuron pairs. In this setup, neurons are arranged into pairs $(a,b)$, where neuron $a$ fires first and neuron $b$ follows after an interval $\delta$. This enforces a directional temporal dependency that cannot be reduced to firing rate or single-neuron ISI statistics.

Formally, each sample contains $N$ neurons, where $N$ is even, grouped into fixed disjoint pairs $(2i, 2i+1)$ for $i = 0, \dots, N/2 - 1$. For a firing rate $r$ (Hz) and total duration $T$ (ms), the number of spike pairs is $r \cdot \tfrac{2T}{1000}$, and we use the same duration $T$ and range of values of $r$ as in the previous section.

For each pair $(a,b)$, neuron $a$ emits a spike at time $t$, and neuron $b$ is constrained to fire at $t+\delta$, with the additional constraint that no two spike pairs overlap within the interval $[t-\delta,\,t+\delta]$ to avoid conflicts.

\begin{figure*}[!htbp]
  \centering
  \includegraphics[width=\linewidth]{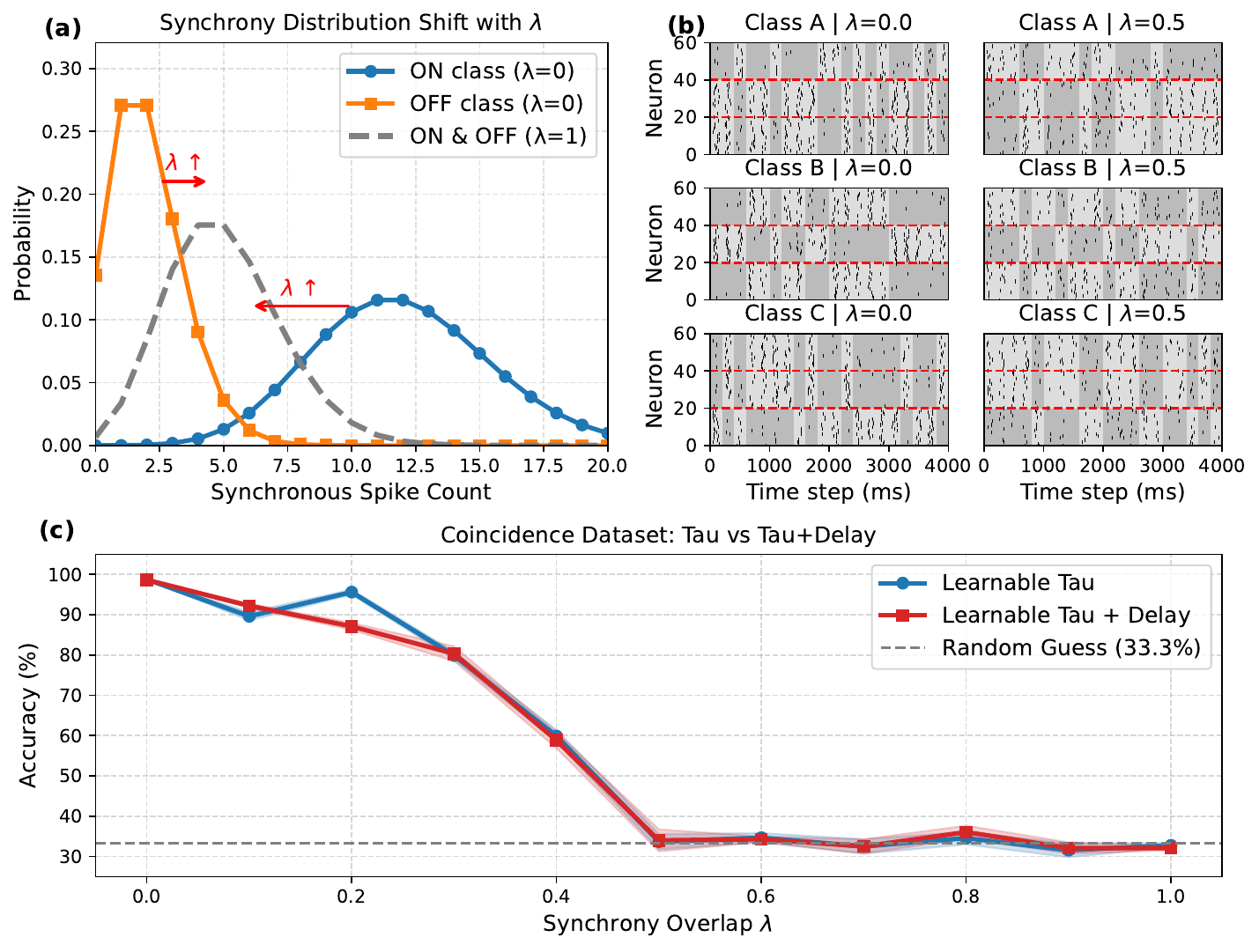}
  \caption{\textbf{Coincidence-based dataset.} 
(a) Spike count distributions for ON/OFF windows at $\lambda=0$ and $\lambda=1$. As $\lambda$ increases, ON and OFF distributions converge.
(b) Spike raster plots for each class under low ($\lambda=0$) and moderate ($\lambda=0.5$) synchrony overlap. Red lines mark group boundaries. 
Light gray and dark gray bands indicate ON and OFF neuron groups, respectively.
(c) Test accuracy across varying synchrony overlap $\lambda$, comparing models with a learnable $\tau$ versus $\tau$+delay.
}
  \label{fig:coin_result}
\end{figure*}

\paragraph{{Results.}}
In the same setup as the previous section, we get similar results for CCISI as we did for the ISI task (\cref{fig:isi_ccisi_result}(d)). However, we find that when we increase the maximum CCISI from 50~ms to up to 500~ms strong differences emerge between these tasks and the two models (Learnable Tau and Tau + Delay). For the ISI task, both models perform similarly over this range (\cref{fig:isi_ccisi_result}(e)). For CCISI however, the difference between the two models becomes pronounced at longer intervals.

This difference between the models with and without delays at longer intervals may be a result of the fact that the membrane time constant is more constrained in having to be long enough to integrate information across all timescales that are present, but short enough to be sensitive to differences on those timescales. By contrast, delays are not so constrained, and are therefore easier to adapt to tasks in which a wider range of timescales are present. We will return to this in \cref{sec:sg_slayer_shd}. {This distinction between integration timescales and explicit temporal alignment is consistent with prior work showing that learnable axonal delays provide a complementary mechanism to heterogeneous membrane time constants for temporal coding \citep{sun2023learnable, sun2025exploiting}.}

\subsection{Surrogate GD Enables Coincidence-Based Temporal Code Learning}
\label{sec:sg_learns_coin}

\paragraph{{Task setup.}}
We next turn to the question of whether or not surrogate gradient descent can learn to extract information encoded in synchrony statistics (rather than precise spike patterns).

In this task, each sample consists of $N = 60$ input neurons evenly divided into three groups (\cref{fig:coin_result}(b)). Time is divided into windows of length $w=200$ time steps, and in each window a group is either ON (high mean firing rate $\mu_\text{on}$) or OFF (low mean firing rate $\mu_\text{off}$). 
The class identity is defined by which two of the three groups fire with synchronous patterns (in every window, both are ON or both OFF) while the third inverts (OFF when the other two groups are ON, and vice versa). For example, Class A corresponds to synchrony between Groups 1 and 2, with Group 3 out of synchrony. To eliminate rate-based cues, ON/OFF group activity toggles randomly in each window, and so the total spike count is balanced across groups and samples, meaning that firing rate gives no information about class.
Spike counts are Poisson distributed, and spikes are randomly allocated to times in the central 50\% of each window.
To control task difficulty, we interpolate between ON/OFF spike count distributions via a synchrony overlap factor $\lambda \in [0, 1]$:
\begin{equation}
\begin{aligned}
\mu_{\text{on}} &= (1 - \lambda) \cdot \mu^{(0)}_{\text{on}} + \lambda \cdot \mu_{\text{avg}}, \\
\mu_{\text{off}} &= (1 - \lambda) \cdot \mu^{(0)}_{\text{off}} + \lambda \cdot \mu_{\text{avg}},
\end{aligned}
\end{equation}
where $\mu_{\text{on}}^{(0)} = 12$, $\mu_{\text{off}}^{(0)} = 2$, and $\mu_{\text{avg}} = 5$. As $\lambda$ increases, the ON and OFF spike distributions converge, reducing synchrony discriminability (\cref{fig:coin_result}(a)).

\paragraph{{Results.}}
Again, surrogate gradient descent can learn to solve this task with near perfect accuracy in the clean case ($\lambda=0$), and degrades to chance level as $\lambda\rightarrow 1$ (\cref{fig:coin_result}(c)).
There is little to no difference between the models (with and without delays), unsurprisingly given that time differences play no part in this task.

\section{Validation of Spike Timing Learning on Speech Datasets}
\label{sec:sg_slayer_shd}

\begin{figure*}[!htbp]
  \centering
  \includegraphics[width=\linewidth]{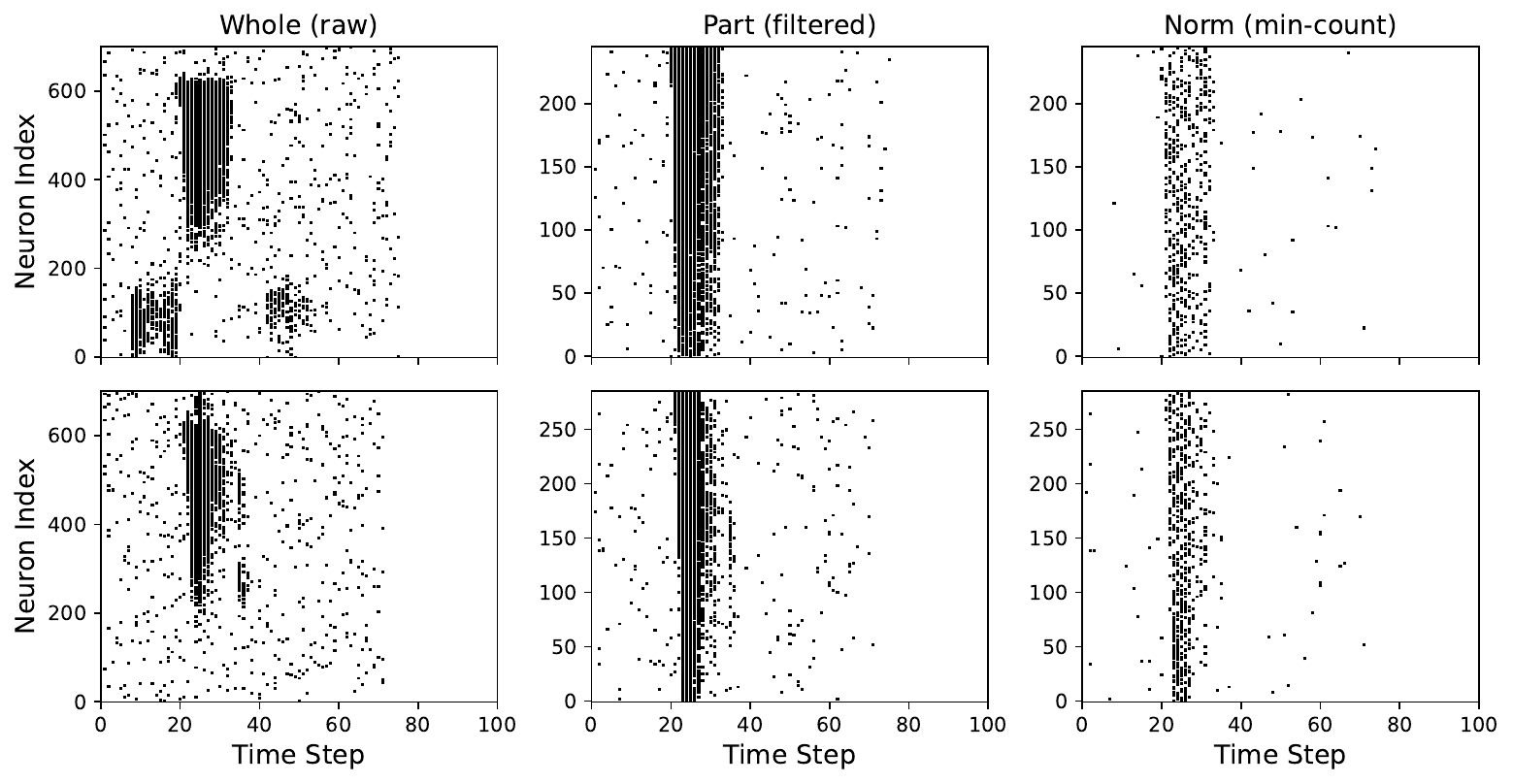}
  \caption{Example spike rasters of auditory datasets SHD (top) and SSC (bottom) across the Whole $\rightarrow$ Part $\rightarrow$ Norm normalization stages. Neuron selection and spike count normalization progressively remove rate-based information while retaining temporal structure. Whole (left) is the original dataset. Part (middle) removes neurons whose spike counts are sometimes very low. Norm (right) randomly selects a fixed number of spikes (different for each neuron) leaving only spike timing information and no rate information.}
  \label{fig:shd_ssc_gen}
\end{figure*}

We have seen that surrogate gradient descent can learn to extract various forms of temporal structure (synchrony and ordered/unordered time differences within/across channels) in controlled conditions.
We now test whether surrogate gradient–trained SNNs can exploit spike timing information in more realistic sensory tasks. 
To this end, we evaluate them on SHD and SSC, which consist of samples of single spoken words from 20/35 classes (respectively) converted to spike trains via a simulated cochlea model  \citep{cramer2020heidelberg}. {Following the standard SHD preprocessing, each sample is discretized into 100 temporal bins \cite{sun2023learnable}.}These datesets exhibit complex spatiotemporal structure but also strong spike-count variation that allows rate-based models to perform surprisingly well. 
To disentangle rate and timing contributions, we construct two timing-normalized dataset variants (\cref{fig:shd_ssc_gen}) that progressively remove rate information while preserving each sample’s temporal structure.

\subsection{Models}
\begin{figure}[!htbp]
  \centering
   \includegraphics[width=0.65\linewidth]{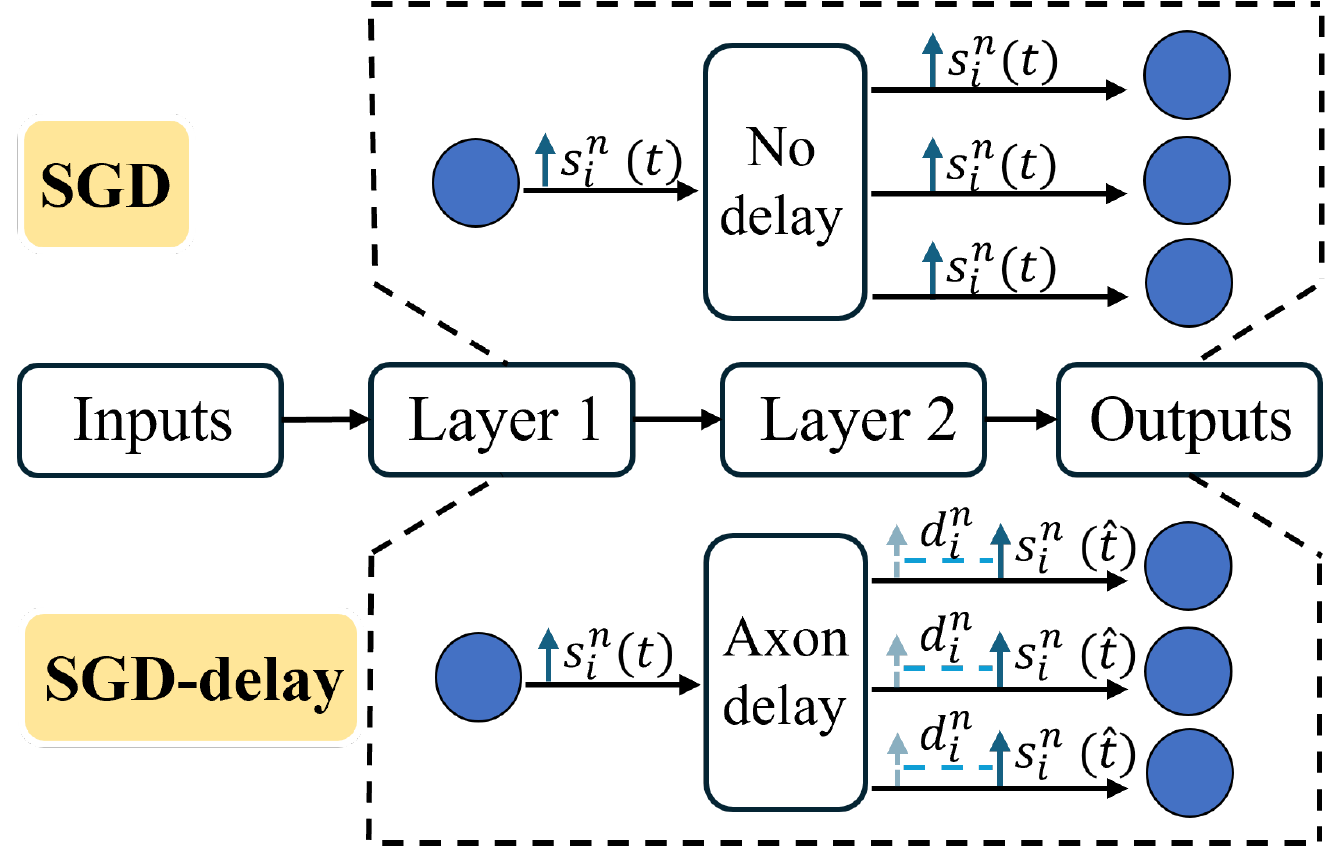}
  \caption{
  {Comparison of Feedforward SNN architectures trained without delays  (SGD, top) and with delays (SGD-delay, bottom). Both models adapt the surrogate gradient descent methods.
  The top row shows a model with no delay components. 
  The bottom row inserts learnable axonal delays between layers. Dashed outlines are used purely as a visual aid to highlight the architectural comparison between the two variants.}
  }
  \label{fig:network_architecture}
\end{figure}

All experiments in this section use a two-layer feedforward SNN (\cref{fig:network_architecture}) implemented in the SLAYER framework \citep{shrestha2018slayer}, with 128 neurons per layer and the Spikemax loss function \citep{shrestha2022spikemax}. 
We compare two variants: a baseline model trained by surrogate gradient descent (SGD), which learns synaptic weights and an enhanced version (SGD-delay) that additionally learns axonal delays \citep{sun2023learnable}. 
Predictions are obtained from the neuron with the highest spike count over the simulation window.

We also compare to the performance of a multilayer perceptron (MLP) using a vector of spike counts for each neuron as input. This serves as a proxy for the performance level attainable using only firing rates, and not spike times. 
The MLP comprises two fully connected layers with {128} hidden units and ReLU activation, mapping the spike count vector to a linear readout over all output classes.

{For the SHD and SSC datasets, we use 200 simulation time steps.
Axonal delays are initialised uniformly in the range $[0,1]$, and an adaptive delay threshold is applied to regulate them during training~\cite{10094768}.
All temporal perturbations (jitter, deletion, reversal) and class-balancing procedures for the Part and Norm variants follow a fixed protocol.}

\subsection{Construction of Timing-Normalized SHD and SSC Datasets}
\label{sec:shd_construction}

While SHD and SSC exhibit complex spatiotemporal spike patterns \citep{cramer2020heidelberg}, spike count vectors vary significantly between classes and consequently, even simple multilayer perceptrons trained only on per‑sample spike counts can achieve nontrivial accuracy (for example, around 50\% for SHD; \cref{fig:shd_ssc_compare}). This makes it hard to determine whether SNNs truly leverage precise timing or simply exploit rate‑based cues.

To address this, we construct two timing-normalized variants of the SHD and SSC datasets, each with an increasing amount of rate-based information stripped away. These variants aim to eliminate spike count confounds while preserving discriminative temporal structure. The normalization process follows a two-stage strategy. {For clarity, we briefly note that the following subsections describe this process in two stages: Section~\ref{sec:shd_construction} first introduces the neuron and sample filtering procedure used to obtain the Part variant, and Section~3.2.2 then explains how the Norm variant is produced by subsampling spike trains to remove all rate-based information while preserving timing.}

\subsubsection{Neuron and Sample Filtering for Spike Count Normalization (Whole $\rightarrow$ Part)}
To eliminate rate-based information while preserving temporal structure, we aim to normalize the spike count for each neuron across all samples. However, some neurons emit zero spikes in certain samples, making exact normalization infeasible. In this first stage, we therefore remove those neurons where the minimum spike count is too low.

Let $X \in \{0,1\}^{M \times N \times T}$ denote the dataset with $M$ samples, $N$ neurons, and $T$ time steps, where $x_{m,i,t} = 1$ indicates a spike from neuron $i$ at time $t$ in sample $m$. For each neuron $i$, we compute its minimum spike count across samples:
\begin{equation}
c_i^{\min} = \min_m \sum_{t=1}^{T} x_{m,i,t}.
\end{equation}

Neurons with $c_i^{\min} < \theta$ (we use $\theta=2$) are initially marked for exclusion. However, in many cases, a neuron's low $c_i^{\min}$ is caused by just a small number of inactive samples. To retain more neurons, we apply an additional sample filtering step: if a neuron's low activity affects fewer than a threshold fraction of samples, we remove those samples instead of the neuron. Formally, for each neuron $i$, we define the set of affected samples:

\begin{equation}
\mathcal{M}_i = \left\{ m \,\middle|\, \sum_{t=1}^{T} x_{m,i,t} < \theta \right\},
\end{equation}
and apply the condition
\begin{equation}
\text{if } \frac{|\mathcal{M}_i|}{M} < \epsilon, \text{ then exclude all samples in } \mathcal{M}_i.
\end{equation}

We set $\epsilon = 0.01$ for both SHD and SSC, reflecting dataset-specific trade-offs. This introduces a balance between neuron retention and sample preservation: deleting more samples allows us to retain more neurons, but excessive deletion can skew the class distribution. To mitigate this, we avoid reducing any class below 50\% of its original size. For example, in SHD, each class initially contains around 500 samples; after filtering, only Class 2 is reduced to 273 samples, while classes like Class 19 are barely affected. This strategy significantly increases neuron retention while preserving sufficient samples per class for training.

After filtering, we retain only neurons that spike at least $\theta$ times in every remaining sample, forming the \textbf{Part} variant.
To ensure a fair comparison across classes, we randomly downsample each class to match the size of the smallest one. This balancing step improves the interpretability of accuracy metrics in the subsequent experiments.

\subsubsection{Min-Count Spike Normalization (Part $\rightarrow$ Norm)}
To fully eliminate spike count information while preserving spike timing, we subsample each spike train to retain exactly $c_i^{\prime \min}$ spikes per neuron $i$ in every sample, where $c_i^{\prime \min}$ is the minimum spike count across the filtered samples defined in the previous stage. Specifically, we randomly select $c_i^{\prime \min}$ spike times from each neuron’s spike train in each sample, ensuring fixed per-neuron spike counts across the dataset. 
This process removes all rate-based information while retaining precise spike timing within each retained spike. However, it may distort global temporal patterns such as inter-spike intervals (ISIs) or synchrony between neurons, depending on which spikes are retained. The resulting \textbf{Norm} dataset preserves the shape of the Part variant and serves as a strict test of spike timing-based learning.

\subsection{Surrogate GD Enables SNNs to Exploit Spike Timing in Realistic Datasets}
\label{sec:shdssc_train}

\begin{figure*}[t]
  \centering
  \includegraphics[width=\linewidth]{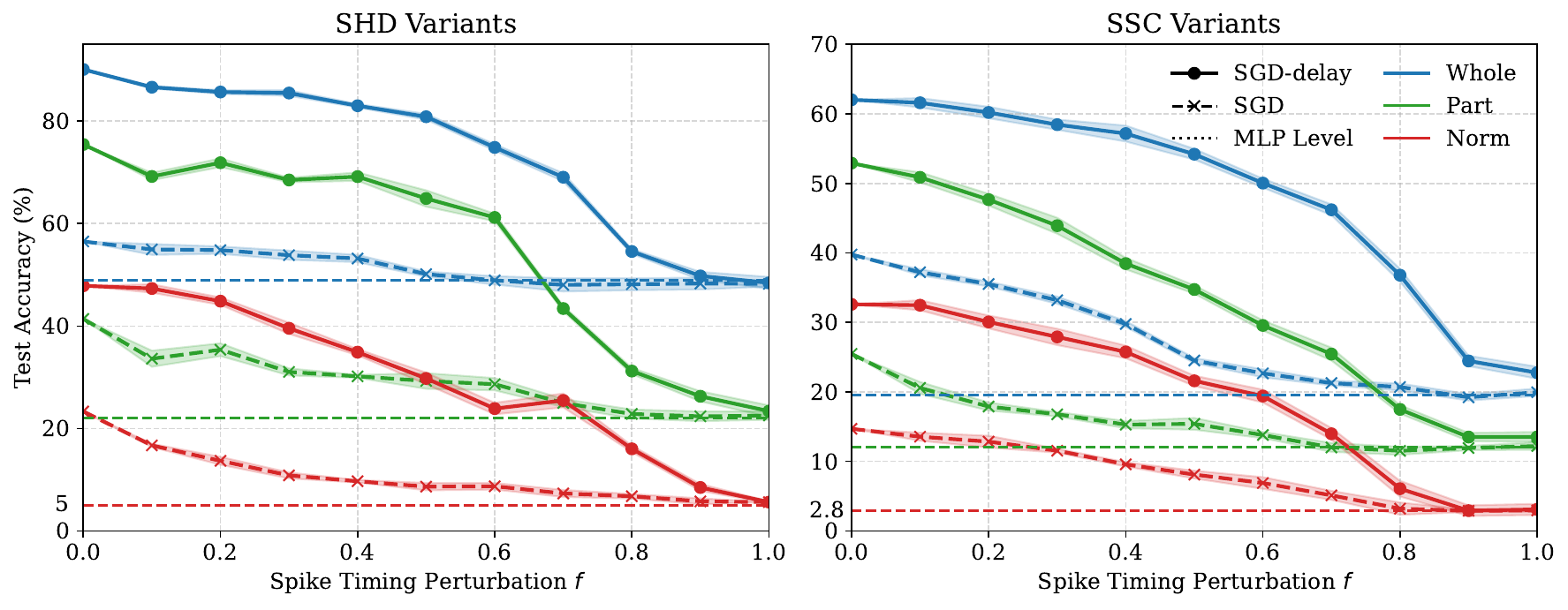}
  \caption{Test accuracy under spike timing perturbation $f$ on SHD (left) and SSC (right) spiking speech datasets for models with (solid lines) and without delays (dashed lines). Performance is shown for the original dataset (whole=blue), on a subset of neurons selected for having a minimum number of spikes (part=green), and on a variant where spike count information has been fully removed (norm=red). For the norm variant, spike count information is fully removed, so the accuracy of an MLP trained on spikes counts (grey dashed) corresponds to chance level (5\% for SHD, 2.8\% for SSC).}
  \label{fig:shd_ssc_compare}
\end{figure*}

We evaluate SGD and SGD-delay on each variant (whole/part/norm), on both the SHD and SSC datasets, across the full range of spike timing perturbations $f\in[0,1]$ (\cref{fig:shd_ssc_compare}). {Here, $f$ is the same perturbation factor used in the synthetic tasks.} {All reported accuracies are averaged over 5 independent runs with different random seeds; shaded regions in the figures indicate mean $\pm$ standard deviation across runs.}
In every case where spike timing information has not been fully removed by perturbation ($f<1$), the spiking neural network model outperforms the MLP trained on the spike counts (which in turn give chance level performance for the Norm variant of each dataset). As $f\rightarrow 1$ the performance of the SNNs approaches the performance of the MLP using firing rates only.
This confirms that surrogate gradient based-learning is not limited to learn rate-based features but can indeed learn from precise spike timing, both with and without explicit delay modeling.

Notably, SGD-delay consistently achieves higher accuracy than SGD across both datasets. For instance, on SHD-norm at $f=0$, SGD achieves 23\% while SGD-delay reaches 48\%; on SSC-norm, the scores are 15\% and 33\% respectively.
Since we have already seen that delays do not substantially improve the ability of SNNs to extract ISI or synchrony, these results suggest that both SHD and SSC datasets contain substantial long cross-channel ISI information. However, we cannot rule out the possibility that the timing information is in another form that we have not considered but that SNNs are able to learn.
These improvements with delays come at a very low parameter cost. The SGD model has $n_in_h+n_h^2+n_hn_o$ parameters (where $n_i$ is the input size which depends on the dataset and variant, $n_h=128$ is the hidden layer size, and $n_o$ is the number of classes which depends on the dataset). The delays in SGD-delay add only $2n_h$ additional parameters. In our case, we increase the number of parameters by at most 0.5\%.

Across both SHD and SSC datasets, we observe a common trend in SGD-delay’s performance: as $f$ increases, the degradation in accuracy accelerates sharply—suggesting strong reliance on high-fidelity spike timing. Around $f{=}0.8$, the curves rapidly drop to the corresponding MLP baseline, after which the decay flattens out.
By contrast, SGD performance without delays has a more steady decline with increasing $f$ suggesting it uses both fewer and less robust spike timing cues.

\section{SNNs With Delays Are More Sensitive to Cross-Channel and Temporal Order Cues}
\label{sec:temporal_reversal}

We have seen that networks using delays seem to be better able to make use of information presented at a range of timescales and involving temporal order / causality (\cref{sec:sg_learns_ccisi}), and that this appears to manifest in better performance in the SHD/SSC datasets (\cref{sec:shdssc_train}). We now probe these insights further with a series of perturbations designed to assess which cues are being used by these models.

\subsection{Models}
\label{sec:temporal_models}

All experiments use the same SNN architecture, training parameters, and loss functions as described in Section~\ref{sec:sg_slayer_shd}. 
{For the jitter and deletion analyses below, new models are trained from scratch under each perturbation setting, motivated by the fact that jitter and deletion are likely present in biological systems at training time. 
In contrast, the time-reversal analysis below reuses the pretrained networks from Section~\ref{sec:sg_slayer_shd}, performing inference on temporally inverted versions of the original datasets, to test out-of-distribution performance. }
Both model variants (SGD and SGD-delay) are evaluated on the SHD and SSC datasets following the same experimental protocol.

\subsection{Biologically Inspired Perturbations}
\label{sec:temporal_perturbation}

To further evaluate the temporal sensitivity of surrogate gradient-trained SNNs, we move beyond the random spike replacement used in earlier sections.

\begin{figure*}[!htbp]
  \centering
  \includegraphics[width=\linewidth]{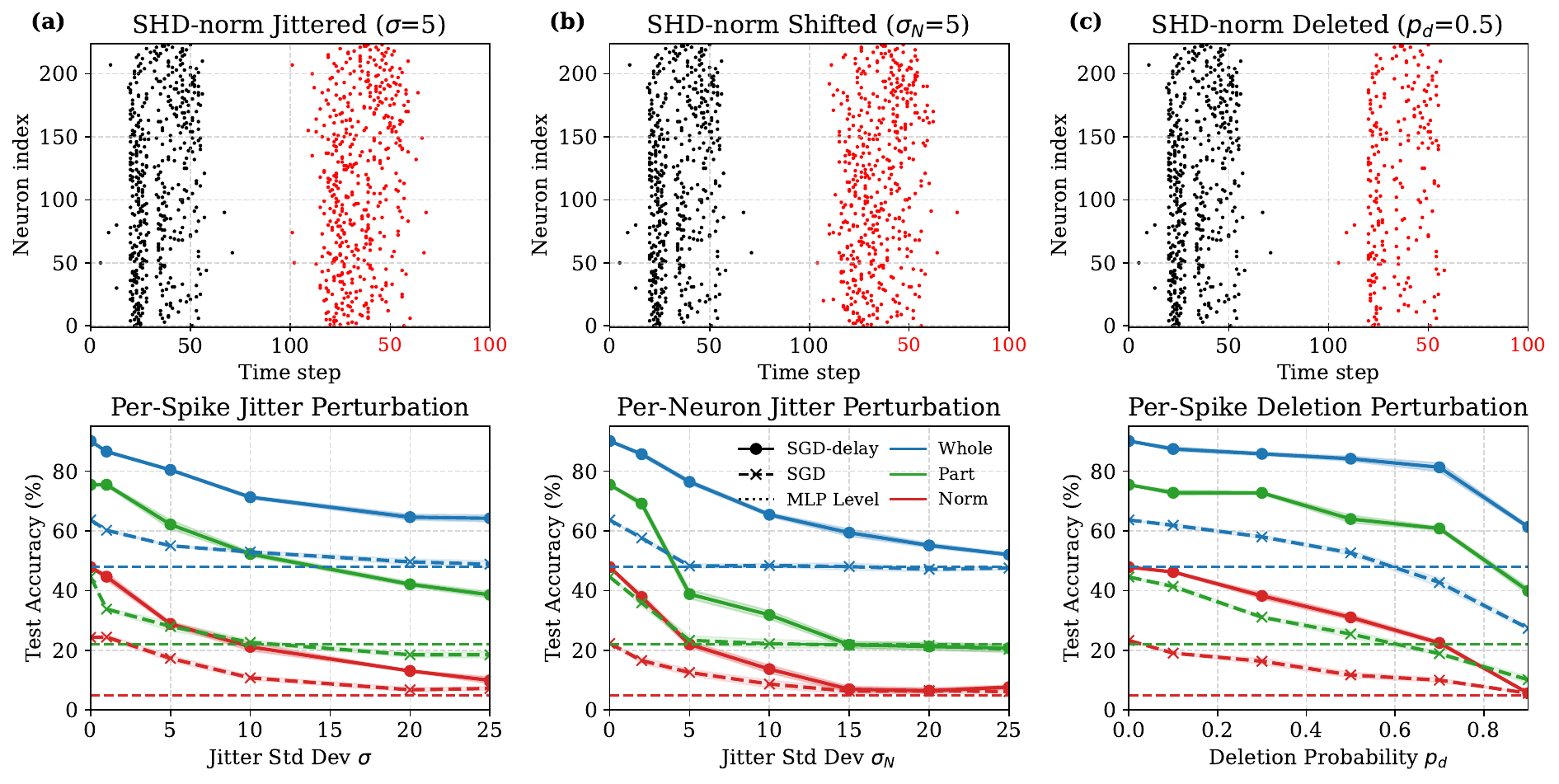}
    \caption{
        Spiking neural network performance under temporal perturbations:
        (a) Per-spike jitter, where each spike is perturbed independently, (b) per-neuron jitter, where all spikes from the same neuron are perturbed equally (c) per-spike deletion. 
        \textbf{Top:} spike rasters of one specific SHD-norm sample before (black) and after (red) perturbation.
        \textbf{Bottom:} test accuracy across three dataset variants (Whole, Part, Norm), using SGD with and without delay learning.
    }

  \label{fig:shd_robustness}
\end{figure*}

First, we look at \textbf{per-spike jitter}, adding a zero-mean Gaussian noise to each input spike time (clipping to the stimulus duration to avoid spike loss). This type of jitter is widely observed in biological neural networks, and we vary the standard deviation $\sigma$ from 0 to 25~ms.

Second, in order to probe the difference between single-neuron timing cues (like ISI) and cross-neuron cues (like CCISI), we also implemented a \textbf{per-neuron jitter}, identical to the per-spike jitter except that the jitter is re-used for all spikes from a given neuron. This type of jitter is less biologically plausible, but leaves single-neuron timing cues unchanged.

Finally, since many spikes fail \emph{in vivo} we study performance at different levels of \textbf{per-spike deletion}, where each spike is independently deleted with probability $p_d$. While the first two types of jitter affect timing but leave spike rate unchanged, this introduces noise for both timing and rate.

\subsection{Delay-Based Networks Use More Cross-Channel Timing Information}

We evaluate model robustness under per-spike jitter, per-neuron jitter and spike deletion on all three SHD variants: whole, part, and norm (\cref{fig:shd_robustness}). First, we compare per-spike versus per-neuron jitter at a range of noise levels ($\sigma, \sigma_N \leq 25$~ms). 
For per-spike jitter, the delay-based model is able to achieve consistently higher performance than the spike rate-only MLP baseline for all noise levels, while the network without delays falls to or below the MLP level (\cref{fig:shd_robustness}(a)). By contrast, for per-neuron jitter, performance falls to the MLP level both with and without delays (\cref{fig:shd_robustness}(c)). Notably, the delay-based model appears much more highly disrupted for the per-neuron jitter than the per-spike jitter. This suggests that the delay-based network is making more use of cross-channel timing information. This is consistent with the differences we saw earlier (\cref{sec:sg_learns_ccisi}), although here the total change to the timing is much smaller. Performance drops significantly by $\sigma_N=10$~ms whereas we only saw big differences for the CCISI task when the maximum ISI was larger than 200~ms. Although not directly comparable, this may suggest that temporal order over small time intervals is a particularly important cue for delay-based networks solving auditory tasks.

Finally, we see that both model variants are fairly robust to spike deletion (\cref{fig:shd_robustness}(c)), suggesting no particular difference in their overall robustness to noise.

\begin{figure*}[!htbp]
  \centering
  \includegraphics[width=\linewidth]{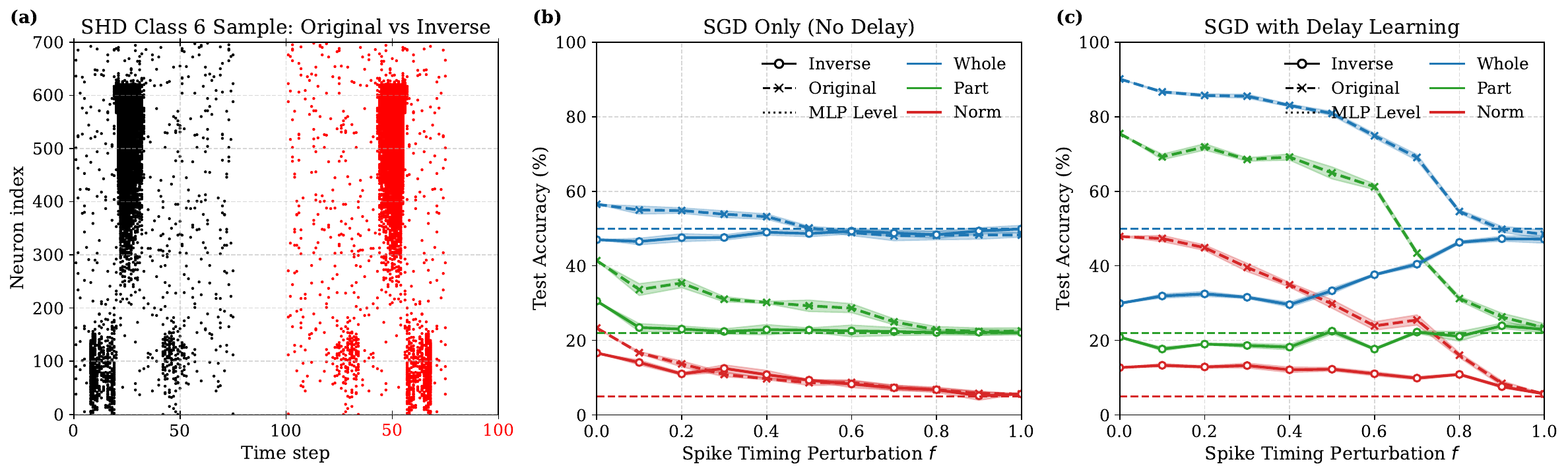}
  \caption{
    {Test accuracy under spike timing perturbation $f$ for original and time-reversed SHD variants. 
    (a) Spike raster of a single sample before and after temporal inversion. 
    (b) Results for models trained without delays. 
    (c) Results for models trained with delays. Each curve corresponds to one of the \textbf{whole}, \textbf{part}, or \textbf{norm} datasets in both original and inverted forms. The MLP level is the performance of a multilayer perceptron trained on spike counts only (no temporal information).}
  }
  \label{fig:inv}
\end{figure*}

\subsection{Delay-Based Networks Are More Sensitive To Time Reversal}

Humans are highly sensitive to the causal structure of sensory input -— e.g., speech played backward becomes unintelligible~\citep{saberi1999cognitive}. To examine whether SNNs exhibit similar temporal sensitivity, we conduct a \textbf{time reversal} experiment by reversing spike trains within their active time windows (\cref{fig:inv}(a)). {Note that in our simple model here, time reversal refers strictly to reversing the temporal order of spikes within each spike train, while preserving spike counts, neuron identities, and the underlying time discretization. A more realistic model of humans listening to time-reversed speed would be to produce new spike trains by running the time-reversed speech through the auditory model, but this would not allow for us to control for keeping the spike counts and ISI distributions identical.}

Given an input $x \in \{0,1\}^{N \times T}$, we identify the minimal interval containing all spikes:
\begin{equation}
[t_{\mathrm{start}},\ t_{\mathrm{end}}] = \left[\min\{t\ |\ x_{i,t} = 1\},\ \max\{t\ |\ x_{i,t} = 1\}\right],
\end{equation}
and reverse each neuron's spike sequence within this interval:
\begin{equation}
x_{i, t}' = x_{i, t_{\mathrm{start}} + t_{\mathrm{end}} - t}, \quad \forall\ t \in [t_{\mathrm{start}},\ t_{\mathrm{end}}],
\label{eq:causal_inversion}
\end{equation}
preserving spike count and neuron identity, but inverting the temporal order.

We train on the original data sets and test on the time-reversed versions. This makes sense when comparing to humans, since we do not listen to a lot of time reversed speech (even if we are big Twin Peaks fans). {It also leaves all single neuron and synchrony timing cues unchanged, only affecting temporal order / causal cues (including CCISI), and therefore allows us to assess the extent to which the models make use of these different types of temporal cues.} We also test at a variety of levels of perturbation $f$ (as in \cref{sec:sg_learns_timing,sec:sg_slayer_shd}).

Our first finding is that the models without delays are much less affected by time reversal than the models trained with delays, across all variants (\cref{fig:inv}(b-c)). This may suggest that delay-based models are a better model of human auditory processing than models without delays.

Our second finding is that for both models, there is a consistent difference between the whole and norm dataset variants. In both cases, under time reversal and no spike timing perturbation ($f=0$), performance is worse than the rate-only MLP baseline. By contrast, for the norm variant, performance still drops but remains better than the MLP baseline. In every case, performance with and without delays matches the MLP performance when all spike timing information is removed ($f=1$). This difference between the whole and norm variants raises a challenge for interpreting these results. We have seen that the whole variant maintains a lot of non-temporal information that is removed by construction in the norm variant. However, it also seems that the whole variant contains much more temporal order structure than the norm variant.

\section{Discussion}\label{sec:discussion}

Surrogate gradient descent and related methods \citep{neftci2019surrogate,shrestha2018slayer} have proven themselves exceptionally capable at the challenging task of training non-differentiable spiking neural networks to carry out challenging tasks.

However, the question remains whether or not the solutions found using these algorithms cover the full range of capabilities of spiking neural networks, or whether they are biased to a particular subfamily of solutions. In both cases, are the solutions found similar to what biological neural networks would learn? We addressed the particular question of the extent to which timing rather than rate of action potentials can be used.

In principle, a timing-based code can be vastly more efficient than a rate-based one.
{This comparison assumes a discretized time representation with finite temporal precision, as used in both neuromorphic datasets and most biological modelling contexts. }
Suppose spike times were accurate to within 1 ms and there was no restriction on firing rate, then for a $k$ ms window, spike count can encode $k+1$ different values whereas with spike timing we can encode $2^k$ values. 
{We emphasize that this comparison is purely illustrative under discretized time assumptions and is not meant to reflect achievable operating regimes in practice.}
In practice, the presence of neural noise and the difficulty of decoding the full range of possible timing codes will limit this. Some of the aspects of spike timing that certainly are robustly decodable include: inter-spike intervals (the time between spikes), cross-neuron coincidences, and at least some more complex spatio-temporal spike patterns. In our results, we have shown that surrogate gradient descent-based methods are able to extract information using all of these aspects of timing-based codes, in both controlled synthetic datasets and messier datasets based on recorded speech. In addition, under various forms of spike time perturbation they degrade gracefully in the majority of cases.

In particular, we looked at the Spiking Heidelberg Digits and Spiking Speech Commands datasets (SHD, SSC). These have become widely used benchmarks in the field of spiking neural networks and neuromorphic computing. We find that these datasets do contain substantial and exploitable temporal information, but that it is also possible to perform surprisingly accurate classification using spike counts only, or following various types of perturbation of the spike timing. As a consequence, this dataset may not be ideal if we wish to probe the extent to which various spike-based models can exploit temporal information. We built derived datasets (SHD-norm and SSC-norm) which, by construction, eliminate all spike count information. (Note that although the only information remaining in this dataset is temporal, it is likely that a significant amount of temporal information is also removed by the transformations applied.) We have made a public release of this dataset to facilitate further research on timing-based computation in spiking neural networks.

Finally, we address the question of whether or not these algorithms find solutions that are biologically feasible by considering what happens when you train a network on a set of sounds, and then test it by time-reversing the spike trains. Since humans perform badly on time-reversed sound, we would expect models that learn in a way that is more similar to humans to also perform badly. We find that spiking neural networks both with and without delays are negatively impacted by time reversal, but that networks with delays are much more significantly impacted.

In future work, it would be interesting to investigate other aspects of spike timing-based codes, for example temporal order (A fires before B), or a more fine-grained analysis of which spatio-temporal spike patterns can be learned. It would also be interesting to devise further datasets of real-world complexity in which temporal information is substantially more important than rate-based information. {Finally, it would also be interesting to understand how and when such networks can generate explicit spike-timing codes at the output.}

\section*{Data and Code}
{{All code in this work is available} on the GitHub repository:  
\url{https://github.com/neural-reckoning/temporal-shd}.  
The datasets are archived on Zenodo\citep{yu2025temporalshd} (\url{https://doi.org/10.5281/zenodo.16153275}).}

\section*{Acknowledgements}\label{sec:acknowledgements}
This project is partly funded by the Advanced Research + Invention Agency (ARIA).
\bibliographystyle{unsrtnat}
\bibliography{paper.bib}

\end{document}